\pdfoutput=1

\documentclass[11pt]{article}

\usepackage[]{conv_ai_ws}
\usepackage{soul}
\sethlcolor{Apricot}
\usepackage{times}
\usepackage{latexsym}

\usepackage[T1]{fontenc}

\usepackage[utf8]{inputenc}

\usepackage{microtype}

\usepackage{hyperref}
\usepackage{url}
\usepackage{wrapfig}
\usepackage{times}
\usepackage{latexsym}
\usepackage{multirow}
\usepackage{xspace}
\usepackage{booktabs}
\usepackage{lineno}
\usepackage{fontawesome}
\newcommand{\atis}{\textsc{ATIS}\xspace}
\newcommand{\snips}{\textsc{Snips}\xspace}

\usepackage{microtype}
\usepackage{graphicx}
\usepackage{amssymb}%
\usepackage{amsmath}
\usepackage{pifont}%
\usepackage{verbatim} %
\usepackage[draft,textsize=footnotesize]{todonotes}

\graphicspath{ {./images/} }

\newcommand\xleftrightarrow[1]{%
  \mathbin{\ooalign{$\,\xrightarrow{#1}$\cr$\xleftarrow{\hphantom{#1}}\,$}}
}

\usepackage{listings}

\definecolor{codegreen}{rgb}{0,0.6,0}
\definecolor{codegray}{rgb}{0.5,0.5,0.5}
\definecolor{codepurple}{rgb}{0.58,0,0.82}
\definecolor{codeyellow}{rgb}{0.67,0.67,0.0}
\definecolor{backcolour}{rgb}{0.95,0.95,0.92}

\lstdefinestyle{mystyle}{
    language=Python,
    backgroundcolor=\color{backcolour},   
    commentstyle=\color{codegreen},
    keywordstyle=\color{magenta},
    emph={testfunc,print,src},
    emphstyle=\color{codeyellow},
    numberstyle=\tiny\color{codegray},
    stringstyle=\color{codepurple},
    breakatwhitespace=false,         
    breaklines=true,                 
    captionpos=b,                    
    keepspaces=true,                 
    numbers=left,                    
    numbersep=5pt,                  
    showspaces=false,                
    showstringspaces=false,
    showtabs=false,                  
    tabsize=2
}

\newcommand{\hlc}[2][yellow]{{%
    \colorlet{foo}{#1}%
    \sethlcolor{foo}\hl{#2}}%
}

\newenvironment{myindentpar}[1]%
 {\begin{list}{}%
         {\setlength{\leftmargin}{#1}}%
         \item[]%
 }
 {\end{list}}

\newcommand{\ignore}[1]{}

\newcommand\blfootnote[1]{%
  \begingroup
  \renewcommand\thefootnote{}\footnote{#1}%
  \addtocounter{footnote}{-1}%
  \endgroup
}

\title{On the Robustness of Intent Classification and Slot Labeling in Goal-oriented Dialog Systems to Real-world Noise}

\author{Sailik Sengupta$^\star$, Jason Krone$^\star$, Saab Mansour\\
\faAmazon~mazon~~\faAmazon WS~~AI\\
\texttt{\{kronej,sailiks,saabm\}@amazon.com}
}

\begin{document}
\maketitle

\blfootnote{$^\star$ Equal Contribution.}
\begin{abstract}
Intent Classification (IC) and Slot Labeling (SL) models, which form the basis of dialogue systems, often encounter noisy data in real-word environments. In this work, we investigate how robust IC/SL models are to noisy data. We collect and publicly release a test-suite for seven common noise types found in production human-to-bot conversations (abbreviations, casing, misspellings, morphological variants, paraphrases, punctuation and synonyms). On this test-suite, we show that common noise types substantially degrade the IC accuracy and SL F1 performance of state-of-the-art BERT-based IC/SL models. 
By leveraging {\em cross-noise robustness transfer} -- training on one noise type to improve robustness on another noise type -- we design aggregate data-augmentation approaches that increase the model performance across all seven noise types by $+10.8\%$ for IC accuracy and $+15$ points for SL F1 on average. To the best of our knowledge, this is the first work to present a single IC/SL model that is robust to a wide range of noise phenomena. 
\end{abstract}

\section{Introduction}
Intent classification (IC) and slot labeling (SL) tasks (shown in \autoref{tab:example_icsl}) form the backbone of goal-oriented dialog systems. In recent times, deep-learning models have achieved impressive performance, reporting accuracies above $95\%$ \cite{chen2019bert} on public IC/SL benchmarks such as \atis~\cite{hemphill1990atis} and \snips~\cite{coucke2018snips}. These datasets, commonly used in the academia, are clean while real-world data is often noisy. Extensive prior work on robustness has focused on identifying noise-types that affect NLP systems and suggesting methods to improve robustness to individual noise types. However, two key questions remain -- (1) what noise types are seen in real-world goal-oriented dialog systems, and (2) how can we construct a single model robust to all real-world noise variants? %

\begin{table}[t]
    \centering
    \small
    \begin{tabular}{ll}
        \toprule
        Input Utterance & \textbf{Play a song by {\hlc[Dandelion]{Bob}} {\hlc[Dandelion]{Dylan}}} \\
        \midrule
        Slot Labels (SL)  & \textbf{O O O O {\color{Dandelion!90!black} artist artist}} \\
        Intent Class (IC)  & \textbf{PlayMusic} \\
        \bottomrule
    \end{tabular}
    \caption{Example input and outputs for IC \& SL tasks.}
    \label{tab:example_icsl}
    \vspace{-0.8em}
\end{table}

In this work, we identify and evaluate the impact of seven noise types frequently observed in a production -- casing variation, misspellings, synonyms, paraphrases, punctuation, abbreviations, and morphological variants -- on IC and SL performance. In this regard, we collect a suite of realistic \atis and \snips test data for these phenomena.
We find that noise reduces the IC/SL performance of state-of-the-art BERT based IC/SL models by an average of $-13.5\%$ for IC accuracy and $-18.9$ points for SL F1. We show that strategic augmentation and regularization approaches can offset these losses, improving performance on noisy benchmarks by $+10.8\%$ for IC and $+15$ points for SL F1 on average by leveraging cross-noise robustness transfer while minimally effecting accuracy on the original test-data. We emphasize that these improvements are for a single IC/SL model that is robust to all noise types. This is a substantial improvement in generalization over existing methods,  which train separate models for each noise type. 

\begin{table*}[t!]
\begin{center}
\footnotesize
\begin{tabular}{ p{2cm}  p{3.2cm}  p{2.5cm}  l }
\toprule
Phenomena & Train Generation & Test Generation & Examples \\[0.15em]
\midrule
Abbreviations & Rule Based & Human & book a flight from San Jose \hl{2} NYC \\[0.15em]
Casing & Rule Based & Rule Based & \hl{BOOK A FLIGHT FROM SAN JOSE TO} NYC \\[0.15em]
Misspellings & \cite{hasan-etal-2015-spelling} & Public DB & book a \hl{flite} from San Jose to NYC \\[0.15em]
Morph. & Internal DB + LM & Human & start \hl{booking} a flight from San Jose to NYC \\[0.15em]
Paraphrasing & Back Translation & Human & \hl{can you book me} a flight from San Jose to NYC \\[0.15em]
Punctuation & \cite{tilk2016} & Human & book a flight from San Jose to NYC\hl{.} \\[0.15em]
Synonyms & WordNet + LM & Human & \hl{reserve} a flight from San Jose to NYC \\
\bottomrule
\end{tabular}
\vspace{-0.39em}
\caption{Summary of methods to generate noisy data for training and testing purposes. As an example, we provide a noised version of the sentence (in ATIS) `book a flight from San Jose to NYC' for each category.}
\label{tab:noise_examples}
\end{center}
\vspace{-1em}
\end{table*}
\vspace{0.8em}
\noindent To summarize our contributions:
\begin{myindentpar}{0.25cm}
    (1) We publicly release a bench-marking suit of IC/SL test data for seven noise types seen in real-world goal oriented dialog systems.
    \footnote{\textbf{Dataset}: 
    \href{https://github.com/amazon-research/real-world-noisy-benchmarks-for-natural-language-understanding}{{\color{blue!60!green}github.com/amazon-research/real-world-noisy-benchmarks-for-natural-language-understanding \faExternalLink}}
    }\\[0.2em]
    (2)
    We show that training data augmentation with one noise type (eg. synonyms) can improve robustness to unseen noise types (eg. paraphrases) substantially (eg. \atis IC: $+12.6$); we call this {\em cross-noise robustness transfer}.\\[0.2em]
    (3) We present a single IC/SL model that improves robustness to all seven noise types that we study by an average of $+10.5\%$ for IC accuracy and $+15$ points for SL F1.\\
\end{myindentpar}

\section{Related Work}
Prior works demonstrate the brittleness of neural systems in NLP to different noise types. Adversarial NLP focuses on construction of examples that impair a model's performance \cite{jia2017adversarial,ribeiro-etal-2018-semantically,ebrahimi2017hotflip,cheng-etal-2019-robust}.
A parallel thread of research investigates the behavior of models towards real-word noise types. Machine Translation (MT) systems perform poorly in the presence of synthetic and natural character-level changes (eg. character swaps, typographical errors, misspellings) \cite{belinkov2017synthetic,karpukhin-etal-2019-training}; similar conclusions hold for casing noise \cite{niu2020robustness}. Other works investigate robustness of NLP models to paraphrases \cite{einolghozati2019improving}, misspellings \cite{pruthi-etal-2019-combating}, and morphological variants \cite{tan-etal-2020-morphin}. The most common approach to improving model robustness is data augmentation, either at pre-training or training time.  
For example, in \cite{tan-etal-2020-morphin}, augmentation with adversarial examples and a single epoch of pre-training can improve robustness of the model for Question Answering (QA) and MT tasks; a comprehensive summary of DA techniques can be found in \cite{feng2021survey}. Recent works present frameworks for comprehensive model-agnostic testing of NLP systems to grammatical and noise variants \cite{ribeiro-etal-2020-beyond,goel2021robustness}. All of these prior works focus on either (1) improving robustness to individual noise types or (2) simply evaluating model performance on synthetically-generated noise variants. In contrast, we construct a single model that is robust to all seven noise types seen in real-world systems. %

Beyond training-data augmentation, pre-training robust word-embeddings is a popular defense against adversarial examples \cite{zhu2019freelb}, misspellings \cite{piktus-etal-2019-misspelling}, and character ordering \cite{malykh-etal-2018-robust}. In  \cite{kudo2018subword}, authors investigate the impact of different sub-word tokenization techniques on model robustness. Effective regularization has also been shown to improve the robustness of IC/SL models to paraphrases \cite{einolghozati2019improving}. 
Yet many of the techniques to improve robustness are not comparable due to evaluation on different test sets. Hence, we provide high-quality noisy test data to serve as a common test bed.

\begin{table}[t!]
\small
\begin{center}
\begin{tabular}{ l  l  c  c  c  c  c }
\toprule
Dataset & {\footnotesize Phenom.} &  \#Utt & \#IC & \#SL & \#SV & BLEU \\
\midrule
\multirow{7}{*}{ATIS} & Original & 893 & 20 & 69 & 288  & 1.00 \\
& Abbrev. & 99 &  13 & 	44 & 	135 & 	0.66 \\
& Case. & 893 & 20 & 69 & 288  & 0.00 \\
&                       Misspl. & 893 & 20 & 69 & 350 & 0.80 \\
&                       Morph. & 115 & 15 & 	45 & 	154 & 0.59	 \\
&                       Para. & 217 &  18 & 	56 & 	185 & 0.42 \\
&                       Punc. & 243 & 14 & 49 & 212 & 0.68 \\
&                       Syn. & 225 &  18 & 	57 & 191 & 	0.64 \\
\cmidrule{1-7}
\multirow{6}{*}{Snips} & Original & 700 & 7 & 39 & 1571 & 1.00 \\
&                         Abbrev. & 98 & 6	 & 35	 & 334	 & 0.63  \\
&                         Case. & 700 & 7 & 39 & 1474 & 0.00 \\
&                         Misspl. & 700 & 7 & 39 & 1623 & 0.83 \\
&                         Morph. & 101 & 6	 & 36	 & 335	 & 0.65  \\
&                         Para. & 197 & 6	 & 36	 & 585	 & 0.54 \\
&                         Syn. & 201 & 6	 & 36	 & 592	 & 0.73 \\
\bottomrule
\end{tabular}
\caption{Statistics on utterance (Utt), intent (IC), slot label (SL), and slot value (SV) counts as well as the average BLEU score between the original and noised utterances for \atis and \snips.} %
\label{tab:datastats}
\vspace{-1em}
\end{center}
\end{table}

\section{Noise Categories}

\looseness=-1
We consider seven types of noise that are prevalent in the traffic of a task oriented dialogue service -- namely, misspellings, casing, synonyms, paraphrases, punctuation, morphological variants, and abbreviations (see \autoref{tab:noise_examples}). With the exception of misspellings and casing, we employ trained data associates to collect test sets that are representative of naturally occurring noise. For the misspelling and casing phenomena, we automatically generate our test sets because (1) high-quality generation is possible for English and (2) purposeful introduction of misspellings is not a natural task. Generating noisy data using human experts is expensive. Therefore, it is not suitable for the purpose of training augmentation. Thus, all of our training augmentation experiments rely upon automatically generated phenomena. In this section, we provide details about the manual data collection process and describe the automatic generation procedure leveraged for training data augmentation. In
\autoref{tab:datastats}, we showcase statistics for \atis and \snips noisy test sets.

\subsection{Manual Test Data Collection}
We employ trained data associates to introduce synonyms, paraphrases, morphological variants, and abbreviations in ATIS and SNIPS. We also use data associates to punctuate ATIS. ATIS requires special treatment for punctuation as it is unpunctuated, while SNIPS is punctuated and does not require manual collection for the punctuation noise type.
Due to cost constraints, we noise a subset of approximately 200 utterances for each dataset.
We instruct the data associates to introduce the given phenomena into the carrier phrase portion of each utterance (i.e. the spans that do not correspond to slot values). In instances where the associates are 
unable to come up with a viable modification to an utterance, the utterance is excluded from the evaluation set. 
For this reason, our test sets for some phenomena, namely abbreviations and morphological variants, contain fewer than 200 utterances. We perform quality assurance on the collected data, using internal data specialist that ensure at least 95\% of the examples in a sample containing 25\% of each noisy test set are realistic and representative of the given noise type.

\subsection{Synthetic Data Generation}
\label{ss:sdg}

\paragraph{Casing}
Casing variation is common in text modality human-to-bot conversations. Consider for example, the responses ``john'' vs. ``John'' vs. ``JOHN'' to a bot prompt ``Can I have your first name?''. Given that the training data is mostly small-cased (for ATIS) and true-cased (for SNIPS), we evaluate the impact of capitalizing all characters in the test set; we simply capitalize all test utterances. For training augmentation, we inject all-caps noise into $50\%$ of training tokens.

\paragraph{Misspellings}
Misspelling test sets contain 15\% misspelled words, sourced from a collection of public, human misspelling-correction pairs\footnote{\href{https://www.dcs.bbk.ac.uk/~ROGER/corpora.html}{{\color{blue!60!green} www.dcs.bbk.ac.uk/~ROGER/corpora.html \faExternalLink}}}. Using human misspelling pairs produces a more natural test set, but it does not generalize well to new languages or domains. Thus, for augmentation, we utilize a probabilistic approach to generate synthetic misspellings put forward in \cite{hasan-etal-2015-spelling}. This method introduces errors based on induction probabilities mined from natural data and bases character level edits on the QWERTY keyboard layout. %

\paragraph{Synonyms}
We augment the training data with synonyms using a two step approach. First, we obtain a list of candidate synonyms via word net. Second, we introduce the synonym candidate that results in the lowest perplexity, as measured by DistilGPT2~\cite{sanh2020distilbert}, into the utterance. This perplexity filter ensures greater fluency. %

\paragraph{Paraphrases}
For training data augmentation, we considered using paraphrased generated using back-translation, from English to Chinese and back~\cite{mallinson2017paraphrasing,einolghozati2019improving}. For preserving slot-labels, we annotate the slot values in the back-translated text if they match the slot values in the original text, ignoring casing differences. In contrast to other noise types that are injected at the token-level, this is injected at the utterance level.

\paragraph{Punctuation}
For ATIS, we augment the training data by using a bi-directional recurrent neural network with attention proposed in \cite{tilk2016}. We note that the punctuation added to ATIS test sets are clearly noise or {\em irrelevant punctuation} \cite{ek2020does} as the absence of these in the original test-sets does not impact intent or slot value classification. While some punctuation can be considered for complex scenarios in NLU (eg. use of comma to separate consecutive slot values), the idea of {\em relevant punctuation} is currently not well explored in the context of NLU systems. We do not experiment with the punctuation phenomena on the Snips dataset, since snips already contains punctuation in both the training and test splits. 

\paragraph{Morphological Variants}
We generate morphological variants for training data augmentation in a similar manner to synonyms. Namely, we first produce a list of candidate morphological variants, using an internal curated database of 120,000 morphological variant pairs sourced by human experts and then select the candidate with the lowest perplexity. While this is similar to the MORPHEUS approach described in \cite{tan2020s}, it has two major differences. First, the list of morphed candidates in our case are chosen based on an internal curated database as opposed to the LemmInflect package\footnote{\href{https://github.com/bjascob/LemmInflect}{\color{blue!60!green} github.com/bjascob/LemmInflect \faExternalLink}}. Second, our approach is a model-agnostic approach that can work regardless of how the model is trained, whereas \cite{tan2020s} uses a model-in-the-loop approach to generate morphed inputs that are adversarial in nature (i.e. flip the model's prediction).

\paragraph{Abbreviations}
To synthetically construct abbreviations for augmenting the training data, we consider first a knowledge base of common abbreviations \cite{beal_2021} and follow certain rule-based approaches to drop vowels from tokens (eg. `people` $\rightarrow$ `ppl') or include common abbreviation mappings between numerals and words which share phonetics (e.g., "2" and "to").

\section{Approach}

In this section, we highlight the various approaches we consider to construct a single model robust to various noise types seen in live traffic.

\subsection{Model Setup}

We evaluate the robustness of a BERT based joint intent classification and slot labeling model, which is currently SOTA on the Snips and ATIS benchmarks~\cite{chen2019bert}.
Similar to \cite{chen2019bert}, we add an additional feed forward layers on top of the [CLS] and sub-token hidden representations to predict the IC and SL tags, respectively.\footnote{Implemented using the gluon tutorial: \href{https://nlp.gluon.ai/model_zoo/intent_cls_slot_labeling/index.html}{\color{blue!60!green} nlp.gluon.ai/model\_zoo/intent\_cls\_slot\_labeling/index.html}} We use the cased BERT checkpoint pre-trained on the Books and Wikipedia corpora.

\subsection{Data Augmentation}

\paragraph{Training Data Augmentation (DA)} For each noise type, we consider augmenting the training data with noised versions of utterances in the original training set along-with the same intent and similar slot labels. For this purpose, we leverage the synthetic approaches described in \autoref{ss:sdg}. %

\paragraph{Aggregate Data Augmentation (A-DA)}
We explore two augmentation strategies that consider all noise types: {\em Uniform (Uni) Aggregation} and {\em Best Proportion (BP) Aggregation}. {\em Uni} augments the training set with $10\%$ for each noise type (i.e. $10\%$ abbreviation, $10\%$ casing, $10\%$ synonyms, etc.). {\em BP} tunes the training augmentation proportions based on the best performance observed on a validation set ($15\%$ abbreviation, $50\%$ casing, $10\%$ synonyms, etc.).

\subsection{Regularization}

\paragraph{BERT-based Sub-word Regularization (B-SR)}
There exists multiple options when splitting word tokens into sub-words. In \cite{kudo2018subword}, authors propose stochastic sub-word tokenization and show that this can improve the robustness of a Machine Translation system.
Given we use a pre-trained BERT model with a fixed vocabulary, we consider an adaptation of this method-- we first consider an index-based split of each word followed by BERT-based tokenization of each split. Then, we sample each such token with $\frac{1}{4}$-th the probability of the original word tokenization. We call this method BERT-based Sub-word Tokenization (B-ST). For example, consider the word `fly', the subwords and the sampling probabilities are as follows. 
\begin{center}
\small
    \begin{tabular}{ccc}
    \toprule
    Manual Splits & BERT sub-word(s) & Sampling Pr \\
    \midrule
    {\tt None}  & `fly' & $0.666$ \\
   `f',`ly'  &  `f', `ly' & $0.167~(0.666/4)$ \\
   `fl',`y'  & `f', `l`, `y' & $0.167~(0.666/4)$\\ 
    \bottomrule
    \end{tabular}
\end{center}

\paragraph{Adversarial Logit Pairing (ALP)} In \cite{einolghozati2019improving}, authors proposed the use of data augmentation (data generated via back-translation) along with regularization terms that penalize (via $L_2$ distance) the model for outputting different intent and slot labeling logits for a training sample and its noise-augmented version. We adapt this approach for our setting, where the A-DA training set is augmented with multiple noise types. %
For each mini-batch with $K_I$ pairs of samples derived from an input utterance $x$ and $K_S$ pairs of slot values, the loss function looks as follows,%
\begin{align}
\scriptstyle
    \mathcal{L} = \frac{1}{K_I} \sum_{i=0}^{K_{I}} ||I(x) - I(\tilde{x})|| + \frac{1}{K_S} \sum_{i=0}^{K_{S}} ||S(x) - S(\tilde{x})|| \nonumber
\end{align}%
where $I$ and $S$ represents the intent and slot labeling logits respectively.

\begin{table*}[t]
\footnotesize
\begin{center}
\begin{tabular}{ l c c | c c | c c | c c}
\toprule
& \multicolumn{4}{c}{ATIS} & \multicolumn{4}{c}{Snips} \\
\multirow{2}{*}{Model} & \multicolumn{2}{c}{IC Acc.} & \multicolumn{2}{c}{SL F1} & \multicolumn{2}{c}{IC Acc.} & \multicolumn{2}{c}{SL F1} \\ 
& Orig. & Noisy & Orig. & Noisy & Orig. & Noisy & Orig. & Noisy \\
\midrule
BERT   & {98.8}         &	{97.3}          &	{95.6}                   &	{86.5}                   &	99.0 &	{98.7}                   & 	96.3 &	{91.2} \\
\cmidrule{2-9}
\hspace{3mm}+MLM Aug$.^{5\%}$ & \textbf{98.7} & {97.4} & \textbf{95.6} & 86.8 & 98.8 & 98.4 & \textbf{96.6} & {91.9} \\

\hspace{3mm}+Train Aug$.^{20\%}$ &  \textbf{98.7}                &	{\textbf{97.8}} &	95.5          &	{\textbf{94.6}}         &	\textbf{99.1}          &	{\textbf{98.8}}          &	96.4          &	{\textbf{95.3}} \\

\hspace{3mm}+MLM \& Train Aug. & 98.6 &  {97.6} & 95.5 & {94.5} & 99.0 & \textbf{98.8} & 96.0 & {95.1}\\[0.2em]
\bottomrule
\end{tabular}
\caption{Robustness of BERT with and without augmentation (Aug.) of misspellings at pre-training time (MLM) and IC/SL training time (Train.) to misspelling phenomena in the ATIS and Snips datasets measured by intent classification accuracy (IC Acc.) and slot labeling F1 (SL F1) scores.}
\label{tab:misspelling_res}
\end{center}
\vspace{-1.2em}
\end{table*}

\section{Experiments}
We first evaluate the trade-off between use of data augmentation during pre-training vs. fine-tuning time. We focus the majority of our analysis on fine-tuning time augmentation as we show it outperforms pre-training augmentation.

\begin{table*}[t]
\tiny
\centering
{\atis~}
\begin{tabular}{ l p{0.35cm} p{0.25cm} p{0.25cm} p{0.25cm} p{0.25cm} p{0.25cm} p{0.25cm} p{0.25cm} p{0.35cm} | p{0.35cm} p{0.25cm} p{0.25cm} p{0.25cm} p{0.25cm} p{0.25cm} p{0.25cm} p{0.25cm} p{0.35cm}} 
\toprule
\multirow{2}{*}{{\footnotesize Model}} & \multicolumn{9}{c}{IC Accuracy} & \multicolumn{9}{c}{SL F1} \\
& \cellcolor{LimeGreen!20!white}orig.$^E$ & \cellcolor{Red!15!white}abv. & \cellcolor{Red!15!white}case & \cellcolor{Red!15!white}spl. & \cellcolor{Red!15!white}mor. & \cellcolor{Red!15!white}para. & \cellcolor{Red!15!white}punc. & \cellcolor{Red!15!white}syn. & \cellcolor{ProcessBlue!15!white}gain$^E$ & \cellcolor{LimeGreen!20!white}orig.$^E$ & \cellcolor{Red!15!white}abv. & \cellcolor{Red!15!white}case & \cellcolor{Red!15!white}spl. & \cellcolor{Red!15!white}mor. & \cellcolor{Red!15!white}para. & \cellcolor{Red!15!white}punc. & \cellcolor{Red!15!white}syn. & \cellcolor{ProcessBlue!15!white}gain$^E$ \\
\cmidrule{2-19}
\textbf{BERT IC/SL} & 97.9 & 85.5 & 72.2 & 97.4 & \textbf{97.4} & 77.8 & 98.9 & 76.8 & ~0.0 & 94.5 & 78.0 & 22.7 & 86.5 & 94.7 & 89.8 & 87.8 & 91.0 & ~0.0 \\[0.4em]
\hspace{3mm}\textbf{+B-SR} & 95.7 & 84.8 & 74.7 & 97.6 & 96.2 & 87.2 & 98.5 & 87.8 & +2.6 & 92.8 & 83.6 & 38.5 & 92.7 & 93.1 & 87.7 & 84.1 & 88.4 & +2.3 \\[0.4em]
\hspace{3mm}\textbf{+DA} & & & & & & & & & & & & & & & & & & \\
\hspace{5mm}abbrev. (15\%) & 97.8 & \cellcolor{Yellow!40!white}88.2 & 72.7 & 97.3 & 97.1 & 78.1 & \textbf{99.2} & 78.1 & +0.7 & 94.1 & \cellcolor{Yellow!40!white}90.8 & 36.2 & 90.0 & 94.7 & 89.9 & 85.7 & \textbf{91.6} & +4.0 \\
\hspace{5mm}casing (50\%) & 97.2 & 84.5 & \cellcolor{Yellow!40!white}98.2 & 97.0 & 95.4 & 83.6 & \textbf{99.2} & 82.3 & +4.8 & 93.4 & 81.1 & \cellcolor{Yellow!40!white}\textbf{94.9} & 88.4 & 92.9 & 88.5 & 85.8 & 89.8 & +9.9 \\
\hspace{5mm}misspl. (20\%) & 97.4 & 86.2 & 75.4 & \cellcolor{Yellow!40!white}97.6 & 96.5 & 84.1 & \textbf{99.2} & 81.7 & +2.0 & 94.5 & 87.8 & 50.1 & \cellcolor{Yellow!40!white}94.6 & 94.9 & 89.4 & 87.1 & 90.7 & +6.3 \\
\hspace{5mm}morph. (10\%) & 97.6 & 85.5 & 72.2 & 97.2 & \cellcolor{Yellow!40!white}96.2 & 77.5 & \textbf{99.2} & 75.7 & -0.4 & 94.6 & 77.9 & 18.6 & 86.9 & \cellcolor{Yellow!40!white}94.8 & 89.5 & 89.1 & 91.0 & -0.4 \\
\hspace{5mm}para. (15\%) & 97.3 & 85.2 & 70.8 & 97.1 & 96.5 & \cellcolor{Yellow!40!white}\textbf{91.4} & 98.1 & \textbf{91.8} & +3.5 & 93.2 & 75.6 & 27.3 & 85.9 & 93.9 & \cellcolor{Yellow!40!white}88.8 & 93.1 & 90.1 & +0.4 \\
\hspace{5mm}punc. (15\%) & 97.8 & 86.9 & 67.9 & 96.8 & 96.8 & 87.5 & \cellcolor{Yellow!40!white}\textbf{99.2} & 85.6 & +2.0 & 93.1 & 77.2 & 14.3 & 85.4 & 93.8 & 88.2 & \cellcolor{Yellow!40!white}93.5 & 89.8 & -1.4 \\
\hspace{5mm}syn. (10\%) & \textbf{97.9} & 88.2 & 72.4 & 97.0 & 96.5 & 91.0 & 98.9 & \cellcolor{Yellow!40!white}90.5 & +4.1 & \textbf{94.8} & 78.3 & 24.2 & 87.7 & 94.8 & \textbf{90.4} & 89.5 & \cellcolor{Yellow!40!white}91.1 & +0.8 \\[0.4em]
\hspace{3mm}\textbf{+A-DA} & & & & & & & & & & & & & & & & & & \\
\hspace{5mm}Uniform (Uni) & 97.3 & 87.5 & 89.5 & 97.5 & 95.9 & 89.8 & \textbf{99.2} & 90.2 & +6.1 & 94.5 & 90.6 & 85.3 & 93.3 & \textbf{95.0} & 89.6 & 93.9 & 90.8 & +12.6 \\
\hspace{5mm}Best Proportion (BP) & 97.5 & 89.2 & \textbf{98.3} & \textbf{98.0} & 97.1 & \textbf{91.4} & \textbf{99.2} & 91.7 & \textbf{+8.3} & 94.5 & \textbf{92.0} & \textbf{94.9} & \textbf{95.0} & \textbf{95.0} & 89.4 & \textbf{94.4} & 90.5 & \textbf{+14.4} \\[0.4em]
\hspace{3mm}\textbf{+A-DA + ALP} & 97.0 & 86.5 & 89.2 & 97.2 & 95.1 & 91.0 & 98.9 & 90.2 & +5.9 & 94.2 & 88.0 & 84.2 & 92.8 & 94.7 & 89.2 & 93.3 & 91.5 & +11.9 \\[0.4em]
\hspace{3mm}\textbf{+A-DA + B-SR} & 97.3 & \textbf{91.6} & 97.9 & 97.5 & 96.5 & 89.3 & \textbf{99.2} & 88.4 & +7.7 & 93.8 & 90.9 & 94.5 & 94.6 & 93.8 & 89.4 & 94.0 & 89.7 & +13.7 \\[0.4em]
\hspace{3mm}\textbf{+A-DA + ALP + B-SR} & 96.8 & 89.2 & 97.7 & 97.3 & 95.9 & 89.9 & 98.9 & 89.3 & +7.3 & 93.5 & 90.4 & 94.3 & 94.4 & 93.5 & 89.3 & 94.0 & 88.8 & +13.3 \\[0.75em]
\end{tabular}

{\snips~}
\begin{tabular}{ l p{0.35cm} p{0.25cm} p{0.25cm} p{0.25cm} p{0.25cm} p{0.25cm} p{0.25cm} p{0.25cm} p{0.35cm} | p{0.35cm} p{0.25cm} p{0.25cm} p{0.25cm} p{0.25cm} p{0.25cm} p{0.25cm} p{0.25cm} p{0.35cm}} 
& \cellcolor{LimeGreen!20!white}orig.$^E$ & \cellcolor{Red!15!white}abv. & \cellcolor{Red!15!white}case & \cellcolor{Red!15!white}spl. & \cellcolor{Red!15!white}mor. & \cellcolor{Red!15!white}para. &
\cellcolor{Red!15!white}punc. &
\cellcolor{Red!15!white}syn. & \cellcolor{ProcessBlue!15!white}gain$^E$ & \cellcolor{LimeGreen!20!white}orig.$^E$ & \cellcolor{Red!15!white}abv. & \cellcolor{Red!15!white}case & \cellcolor{Red!15!white}spl. & \cellcolor{Red!15!white}mor. & 
\cellcolor{Red!15!white}para. &
\cellcolor{Red!15!white}punc. &
\cellcolor{Red!15!white}syn. & \cellcolor{ProcessBlue!15!white}gain$^E$ \\
\cmidrule{2-19}
\textbf{BERT IC/SL} & 98.3 & 97.6 & 23.3 & 98.7 & 98.0 & 98.5 & $-$ & 98.5 & ~0.0 & \textbf{97.0} & 88.1 & 5.0 & 91.2 & 96.5 & 94.9 & $-$ & \textbf{96.3} & ~0.0 \\[0.4em]
\hspace{3mm}\textbf{+B-SR} & 98.2 & 98.3 & 20.2 & 98.4 & 98.0 & 96.8 & &  98.3 & -0.8 & 93.9 & 84.6 & 3.0 & 92.2 & 91.6 & 91.1 & & 92.5 & -3.3 \\[0.4em]
\hspace{3mm}\textbf{+DA} & & & & & & & & & & & & & & & & & & \\
\hspace{5mm}abbrev. (15\%) & 98.3 & \cellcolor{Yellow!40!white}98.3 & 24.7 & 98.1 & 98.3 & 98.1 & $-$ & 98.3 & +0.2 & 96.8 & \cellcolor{Yellow!40!white}\textbf{96.0} & 3.0 & 92.7 & \textbf{97.4} & \textbf{95.2} & $-$ & 96.1 & +1.4 \\
\hspace{5mm}casing (50\%) & 98.6 & 98.0 & \cellcolor{Yellow!40!white}\textbf{98.7} & 98.2 & 98.3 & 98.3 & $-$ & 98.0 & +12.5 & 96.1 & 89.6 & \cellcolor{Yellow!40!white}90.2 & 91.5 & 97.2 & 93.8 & $-$ & 94.3 & +14.0 \\
\hspace{5mm}misspl. (20\%) & 98.7 & 97.3 & 24.6 & \cellcolor{Yellow!40!white}98.9 & 98.3 & 98.8 & $-$ & 98.7 & +0.4 & 96.6 & 91.3 & +5.1 & \cellcolor{Yellow!40!white}\textbf{94.9} & 97.0 & 95.0 & $-$ & 95.9 & +1.1 \\
\hspace{5mm}morph. (10\%) & 98.3 & 98.3 & 19.4 & 98.4 & \cellcolor{Yellow!40!white}98.0 & 98.3 & $-$ & 98.7 & +2.7 & 96.8 & 85.6 & 1.4 & 91.3 & \cellcolor{Yellow!40!white}96.8 & 95.1 & $-$ & 96.0 & -2.7 \\
\hspace{5mm}para. (15\%) & 98.6 & 97.6 & 29.1 & 98.3 & 98.0 & \cellcolor{Yellow!40!white}98.5 & $-$ & 98.2 & +0.9 & 94.3 & 84.6 & 9.2 & 88.1 & 96.0 & \cellcolor{Yellow!40!white}92.5 & $-$ & 93.5 & -1.8 \\
\hspace{5mm}syn. (10\%) & 98.3 & 98.0 & 21.1 & 98.5 & 98.3 & 98.1 & $-$ & \cellcolor{Yellow!40!white}98.8 & -0.3 & 96.7 & 87.5 & 5.5 & 91.2 & 97.2 & 94.2 & $-$ & \cellcolor{Yellow!40!white}95.5 & -0.2 \\[0.4em]
\hspace{3mm}\textbf{+A-DA} & & & & & & & & & & & & & & & & & \\
\hspace{5mm}Uniform (Uni) & 98.6 & 97.6 & 91.2 & \textbf{99.0} & 98.3 & 98.8 & $-$ & 98.3 & +11.5 & 96.4 & 94.3 & 67.1 & 93.9 & 96.7 & 94.7 & $-$ & 95.3 & +11.6 \\
\hspace{5mm}Best Proportion (BP) & 98.8 & 98.6 & 98.6 & 98.9 & \textbf{98.7} & \textbf{99.2} & $-$ & 98.8 & +13.1 & 95.8 & 95.7 & \textbf{90.4} & 94.6 & 96.6 & 94.8 & $-$ & 95.0 & \textbf{+15.6} \\[0.4em]
\hspace{3mm}\textbf{+A-DA + ALP} & 98.9 & 98.6 & 98.6 & 98.7 & \textbf{98.7} & 99.0 & $-$ & 99.2 & +13.1 & 95.8 & 94.8 & 86.5 & 94.6 & 96.6 & 94.8 & $-$ & 94.9 & +14.8 \\[0.4em]
\hspace{3mm}\textbf{+A-DA + B-SR} & \textbf{99.2} & \textbf{99.0} & 98.5 & 98.9 & 98.0 & \textbf{99.2} & $-$ & \textbf{99.3} & \textbf{+13.2} & 95.2 & 94.3 & 89.5 & 94.0 & 94.7 & 93.7 & $-$ & 94.5 & +14.5 \\[0.4em]
\hspace{3mm}\textbf{+A-DA + ALP + B-SR} & 98.8 & 98.6 & 98.5 & 99.0 & 98.3 & 99.0 & $-$ & \textbf{99.3} & +13.1 & 95.0 & 93.5 & 87.9 & 94.1 & 94.5 & 92.9 & $-$ & 94.1 & +13.8 \\
\bottomrule
\end{tabular}
\caption{Intent Classification (IC) and Slot Labeling (SL) metrics for the different approaches on \atis and \snips averaged over $3$ seed runs. The column with the {\hlc[LimeGreen!20!white]{Orig.$^E$}} header reports average accuracy over the different phenomena control sets, the red-highlighted columns showcase results on the noised-sets and the {\hlc[ProcessBlue!15!white]{Gain$^E$}} shows the average gain over the BERT IC/SL baseline across all datasets. The percentage of augmented data for the (DA) methods is shown in brackets. Yellow cells highlight performance of {\hlc[Yellow!40!white]{Noise-Aligned Data Augmentation (NADA)}} approaches. As \snips lacks a punctuation noised-set, the results are omitted for this condition.}
\label{tab:results}
\vspace{-1.1em}
\end{table*}

\subsection{Pre-training {\em vs.} Fine-tuning DA for Synthetic Misspellings}
\label{sec:np_misspell}
We pre-train BERT on the Wikipedia and Books corpus augmented with synthetic misspellings at a rate of $5\%$ for an additional $9,500$ steps using the standard MLM objective. 
Our results, shown in \autoref{tab:misspelling_res}, empirically demonstrate that noisy pre-training does not perform better than training augmentation (esp. on the noisy data). While noisy pre-training generates marginal improvements on SL F1 (avg. $+0.3$), training time augmentation produces substantially larger gains ($+6.1$ on SL F1 and $+0.5\%$ on IC acc). Due to this finding, we only  consider data augmentation at the fine-tuning stage in the experiments that follow.

\subsection{Results}

\autoref{tab:results} highlights the performance of the various methods (represented by the different rows) for IC accuracy and SL F1 on the {\hlc[LimeGreen!20!white]{original test set}} (averaged over the control sets for each noise type), {\hlc[Red!15!white]{treatment sets}} for the individual noise types, and an {\hlc[ProcessBlue!15!white]{average gain}} over the control and treatment sets of each approach over the BERT IC-SL baseline. Figures in \textbf{bold} indicate the best performance for each column.

The diagonal cells under the data augmentation (DA) section, highlighted in {\hlc[Yellow!50!white]{yellow}}, represent the effect of augmentation with a particular noise type and testing on it. While this idea of {\em Noise-Aligned Data Augmentation (NADA)} has been extensively explored in prior work, we observe that only $5$ of the yellow cells (namely, punc. IC, casing SL for ATIS and casing IC, abbrev. SL, misspell SL for Snips) result in the best performing models (out of $26$ metrics on noised-test sets). This indicates that augmentation with other noise types (cross-noise) has beneficial (side-)effects on the robustness to a particular noise type; we term this phenomena as {\em Cross-noise Robustness Transfer}.

We discuss the results of our experiments in three sections. First, we summarize the performance gains of the different models across all the test sets. Then we analyze the performance on original test sets. Lastly, we analyze performance on the noisy test data.

\subsubsection{Overall Performance}

\paragraph{Data-augmentation approaches (DA)}
We find that A-DA with BP aggregation outperforms all the approaches on \atis (IC:$+8.4\%$, SL:$+14.4$) and \snips (IC:$+13.1\%$, SL:$+15.6$). NADA approaches, while often successful for a particular noise type, do not perform well against all noise types and hence, aggregate DA methods are deemed necessary. Further, we observe that injecting a particular noise in training data provides robustness against other noise types. Thus, we attribute the success of the A-DA BP approach to this cross-noise robustness transfer. We find that uniform sampling of noise types for aggregate data-augmentation (A-DA Uni) is always worse than best-proportion aggregation (A-DA BP), whether used by itself or in conjunction with the regularization approaches. Thus, we only showcase the results of using A-DA BP with regularization approaches in \autoref{tab:results}.

\paragraph{BERT-based Sub-word Regularization (B-SR)}
B-SR, by itself, provides marginal gains for \atis (IC:$+2.7\%$, SL:$+2.3$) and degrades the performance of the baseline on \snips (IC:$-0.8\%$, SL:$-3.3$). For \atis, the gains stem from minor improvements seen on paraphrase and synonym test-data; for \snips, we see a consistent degradation in performance for both the original data and all noise types. In contrary, when coupled with A-DA BP approach, it produces the best performing model on \snips IC ($+13.2\%$) marginally edging out A-DA's gain ($+13.1\%$) and a fraction behind the best performing models in other categories.

\paragraph{Adversarial Logit Pairing (ALP)~}
While ALP coupled with A-DA BP boosts performance when compared to the (BERT IC-SL) baseline model on \atis (IC:$+5.9\%$, SL:$+11.9$) and \snips (IC:$+13.1\%$, SL:$+14.8$) data, it does not yield the best-performing model robust to all noise types.

An amalgamation of the approaches, although better than the baseline on \atis (IC:$+7.3\%$, SL:$+13.3$) and \snips (IC:$+13.1\%$, SL:$+13.8$) data, lacks behind the best-performing models. This indicates that the approaches A-DA, ALP, and B-SR don't provide complimentary benefits and thus, cannot simply be combined to construct a more robust model.

\subsubsection{Performance on the Original Test Sets}

\looseness=-1
Owing to the lack of linguistic diversity in the original test-sets, we do not observe models trained on grammatically-correct noise data (eg. paraphrase, morphological variance, synonyms) clearly outperform the joint-BERT baseline.
Yet, synonym-based DA results in the best performing model on both IC and SL performance for \atis. A key reason for this is that \atis has domain-specific vocabulary-- eg. training data maps utterances with the word `flight' to atis\_flight and `plane' to atis\_aircraft even though the two words can be used interchangeably in test utterances. Synonym-based DA helps the model to generalize better by allowing the model to  avoid this spurious co-relations between words and intents and consider the larger context of the sentence/utterance. For \snips, which has a more diverse vocabulary, well distributed over the test and training set compared to \atis, the original BERT IC-SL performs the best on SL F1, whereas the model A-DA (BP) + B-SR performs the best on IC accuracy. The diverse tokenization of words, caused by both A-DA and B-SR, ensures that words in movie/book/song names, which may often be indistinguishable, does not impact the model's IC capabilities and the model pays more attention to the carrier phrases.

\subsubsection{Performance on Noised Test Sets}

For each noise type, we design a control set that contains only the clean version of the noised utterances in the treatment set. While we only list average performance on the original/control test sets in \autoref{tab:results}, our discussion below often analyzes the model performance on the treatment set and compares it to the corresponding control test set (an opposed to only drawing inference based on the average metrics).

\vspace{-0.3em}

\paragraph{Abbreviations~}
\looseness=-1
Abbreviation degrades IC ($-12.34\%$ for \atis, $-0.7\%$ for \snips) and SL performance ($-16.47$ for ATIS, $-8.87$ for \snips). Note that the degradation on \atis is of higher magnitude than on \snips. Upon investigation, we observe that ATIS abbreviations are often specific to the travel domain (e.g., ``tix" vs. ``tickets"), whereas abbreviations for \snips are more general (e.g., ``@" vs. ``at"). We hypothesize that BERT is more likely to have been trained on the abbreviations present seen in the \snips noised set than for \atis, and therefore more robust on \snips.

Augmentation with only abbreviation data boosts performance on SL-F1 but provides small gains on IC accuracy. We find that the coverage of test abbreviations (injected manually) in the augmented training data (injected synthetically) is far lower for \atis than for \snips (40\% vs. 74\%) and this results in a higher disparity between IC performance on clean ($97.86\%$) \textit{vs.} noised ($88.22\%$) test-set for \atis even after augmentation. Further, we observe that augmentation with synonyms also provides a boost in IC accuracy for abbreviation injected test-data for both \atis and \snips. This cross-noise robustness transfer further boosts the accuracy when the model is trained with BP A-DA (that contains $15\%$ abbreviation and $10\%$ synonym data). When further coupled with B-SR, the best performing model achieves an IC accuracy of $91.58\%$ for \atis ($+5.06\%$) and $99.25\%$ ($+0.93\%$) for \snips. For SL, BP performs the best for \atis while training data augmentation with abbreviation performs the best for \snips.

\vspace{-0.3em}

\paragraph{Casing}
IC and SL performances drop substantially on the noised test set because the Bert tokenizer fails to identify fully capitalized words in the vocabulary and instead breaks them down to often match character-level sub-word tokens (eg. `would' $\rightarrow$ \{`W', `O', `U', `LD'\}). Without the use of augmentation, where half-of-the training data is injected with capitalized form of words, the classifier is not able to associate these sub-token representations to the correct intent classes or the slot-labels. Unfortunately, casing noise is orthogonal to all the other noise types (as injection other noise types does not improve model performance on casing test-set) and thus, does not benefit from cross noise-robustness transfer. While the best performing model for \atis (IC: $98.28\%$, SL: $94.88$) and \snips SL ($90.37$) is the BP A-DA, it is only marginally better ($<0.2$) than the training data augmentation approach, which is a better for \snips IC. Note that as the BP A-DA approach considers augmented data will all the noise types, it significantly increases the training time per epoch (see appendix for run-times of the various approaches). Hence, if one desires a model to only be robust to casing noise, we suggest using training data augmentation with capitalized words.

\vspace{-0.3em}

\paragraph{Misspelling}
Test-time misspellings do not impact IC accuracy more than $0.2\%$ points; a misspelled word in the utterances changes the sub-token breakdown of that word (`what' {\em vs.} \{`wa',`t'\}) that, in turn, does not change the intent of the sentence. While majority of the noise in the test set is in the carrier phrase tokens (as opposed to the tokens representing slot values), we notice a large drop in SL F1 (\atis: $-8.6$, \snips: $-5.1$). For \atis, we noted that slot values for slot labels such at period of day (eg. `night' {\em vs.} `nite') are more prone to mis-classification when injected with misspelled noise in comparison to slot values such as day names (`sunday' {\em vs.} `suntday').  Augmentation of training data with $20\%$ misspellings increases SL F1 (\atis: $+8$, \snips: $+3$) making it within $1.2$ points of performance on the control set. As the sub-tokens of misspelled words are now better recognized as slot values. This result demonstrates training on synthetic misspellings \cite{hasan-etal-2015-spelling} can improve generalization to natural misspelling noise seen at test time. Further, we observe that other noise types such as abbreviation, which also increases the model's familiarity to sub-word tokens, also boosts the model's accuracy to test-time misspelling; this cross-noise robustness transfer makes BP A-DA the best performing model for \atis. Given tokenization is a key issue for the performance drop, we observe B-SR provides a boost in model performance, but comparatively less than A-DA approaches. Also, coupling A-DA + B-SR doesn't necessarily boost the accuracy further.

\paragraph{Morphological Variants}
In contrast to the lack of robustness seen on various NLP tasks in the presence of morphological noise \cite{tan2020s}, IC and SL models evaluated on \atis and \snips data are pretty robust to morphological variation at test-time, deviating from test-time performance on clean data by at most $-1.1$ points. We notice that synonyms and paraphrase noise injected at training time helps in improving the models robustness to morphological variance at test time. Upon using A-DA BP, we observe that this cross-noise robustness transfer boosts the model's performance beyond the baseline model's performance on the clean data.

\vspace{-0.3em}

\paragraph{Synonyms}
Synonyms decrease the model's performance on \atis (IC: $-19.8\%$, SL:$-2.8$), while the impact on \snips IC/SL is negligible. We note that the model picks up word to intent correlations (`fare' $\rightarrow$ `atis\_airfare', `flight(s)' $\rightarrow$ `atis\_flight', `plane(s)' $\rightarrow$ `atis\_aircraft') and when these words are replaced with a synonym in the test data, the model misclassfies the intent. For example, changing the word `flight(s)' to the word `plane(s)' where they can be used as synonyms, the model's intent prediction flips from `atis\_flight' to `atis\_aircraft'. We conjecture that this difference in effect size is the result of over-fitting on less diverse \atis carrier phrases. This lack of diversity is evidenced by the fact that ATIS contains roughly half as many unique carrier phrase tokens as Snips ($430$ {\em vs.} $842$), despite having longer utterances on average ($11.3$ {\em vs.} $9.1$ words).  While introducing synonyms into the \atis training data boosts IC ($+13.7\%$) and SL ($+0.7$) performance, the best performing models are the ones augmented with paraphrase data (for \atis IC) and abbreviation data (for \atis SL). In this setting, the benefit of cross-noise robustness transfer is more pronounced where other noise types can boost the accuracy more than the particular noise type observed at test time. This is also due to the fact that data augmentation with synonyms generated via word-net + language models does not align as well with human-injected synonym noise seen at test time.

\vspace{-0.3em}

\paragraph{Paraphrase}
\looseness=-1
Paraphrases lead to a significant drop in \atis performance (IC: $-18.6\%$, SL:$3.0$) and a marginal drop for \snips (IC:$-0.2$, SL:$-1.9$). Similar to the cases of synonyms, we posit that \atis is impacted due to (1) the lack of diverse carrier phrases in the training set and (2) a greater degree of dis-similarity between the original and the paraphrased test-sets (evident from the $0.12$ lower BLEU score compared to \snips). While, training augmentation with back-translated paraphrases yields comparable scores to the baseline system, ALP \cite{einolghozati2019improving} coupled with Uni A-DA yields the best IC accuracy for both \atis and \snips. For SL, cross-noise robustness transfer with synonyms (for \atis) and abbreviation (for \snips) aids the model in achieving the best performance.

\vspace{-0.3em}

\paragraph{Punctuation~}
\looseness=-1
As mentioned earlier, \snips training data already has (is naturally augmented with) punctuation tokens for a sub-set of utterances, we do not consider punctuation as a noise and hence, do not create a test set for \snips.
BERT makes models robust to punctuation-based noise on IC accuracy, but we observe a $8.6$ point difference on SL F1.  This happens due to either the (1) over-prediction of punctuation tokens after a slot value as a slot-value (eg. in an uttearnce `$\dots$ to Pittsburgh .' both `Pittsburgh' and `.' are identified as city names) or (2) missing out slot values when a punctuation is present around it (eg. in an utterance `$\dots$ from Columbus, Ohio to $\dots$', the model fails to recognize Columbus and Ohio as slot values). While augmenting with punctuation data boosts model performance, we also noticed a similar boost in SL F1 when augmenting training data with back-translated paraphrase data because machine translation often automatically introduced punctuation tokens. Hence, when coupled together in the A-DA approaches, we observe the best model performance with BP A-DA on \atis' SL F1.

\section{Conclusion}
In this paper, we show that SOTA BERT based IC/SL models are not robust to many real world noise types found in production. We further demonstrate that cross-noise robustness transfer -- training on one noise type to improve model robustness on another noise type -- yields gains for a number of noise type pairs. Through Aggregate Data Augmentation (A-DA), we leverage cross-noise robustness transfer to improve the model's average performance by $+10.8\%$ on IC and $+15$ points on SL F1. Despite the gains we obtain, our benchmark provides substantial headroom to improve model performance on the abbreviation, synonym, and paraphrase noise types. We hope that our benchmark will support future research in this direction and enable the design of robust IC/SL models for goal-oriented dialogue systems.

\paragraph{Acknowledgements} We would like to thank the anonymous reviewers at multiple venues and the \href{https://aws.amazon.com/lex/}{\color{blue!60!green} Amazon Lex team \faExternalLink} for their insightful feedbacks and constructive suggestions.

\bibliography{anthology,custom}
\bibliographystyle{acl_natbib}

\clearpage

\appendix
\section{Appendix}
\subsection{Hyper Parameter Settings}

We fine-tune BERT for up to $40$ epochs with a batch size of $32$. To prevent over fitting, we use early stopping on the validation loss. We optimize BERT parameters using gluonnlp's bertadam optimizer with a learning rate of 5e-5 and no weight decay. These are the default hyper-parameters provided in the gluon tutorial for intent classification and slot labeling. \footnote{\url{https://nlp.gluon.ai/model_zoo/intent_cls_slot_labeling/index.html}}

\subsection{Hyper Parameter Tuning}

We use the default BERT joint IC/SL hyper-parameters mentioned above in all experiments, for both the baseline or training augmentation approaches. Using these fixed hyper-parameters, we tune the noise rate used for IC/SL training data augmentation. We tune the training data augmentation noise rate in the ranges listed below for each noise type. We select the noise rate that provides the best trade-off between performance on the control and treatment sets. The noise rates used in our final results are shown in the results table as a superscript.\\

\textbf{Noise Rate Search Range by Noise Type:}
\begin{itemize}
\item Casing: \{15\%, 25\%, 50\%, 100\%\}
\item Misspellings: \{10\%, 15\%, 20\%, 25\%, 30\% \}
\item Abbreviations: \{10\%, 15\%, 20\%, 25\%\}
\item Morphological Variants: \{10\%, 20\%, 25\%, 30\% 50\%\}
\item Synonyms: \{10\%, 20\%, 25\%, 30\%, 50\%\}
\item Paraphrases: \{5\%, 10\%, 15\%, 20\%\}
\end{itemize}

\subsection{Compute Environment}
We run all experiments on p3.2xlarge GPU instances using the AWS Deep Learning AMI (Ubuntu 16.04).  

\subsection{Packages Used in Experimentation}
We utilize the following packages to train and evaluate our models as well as generate synthetic noise for training augmentation:

\begin{lstlisting}[style=mystyle, caption=requirements.txt]
nltk==3.3
gluonnlp==0.8.1
mxnet==1.3.0
numpy==1.14.3
scikit-learn==0.23.1
scipy==1.1.0
torch==1.7.1
transformers==4.2.2
seqeval==0.0.12
\end{lstlisting}

\begin{table}[t]
\footnotesize
\begin{center}
    \begin{tabular}{llc}
        \toprule
        Dataset & Model Type & $\approx$Time Taken \\
        \midrule
        \multirow{9}{*}{\atis} & BERT IC-SL & 13m\\
        & + B-SR & 17m \\
        & + Uni A-DA & 22m \\
        & + Uni A-DA + B-SR & 27m \\
        & + Uni A-DA + ALP & 37m \\
        & + Uni A-DA + ALP + B-SR & 58m \\
        & + BP A-DA & 57m \\
        & + BP A-DA + B-SR & 65m\\
        & + BP A-DA + ALP & 125m\\
        & + BP A-DA + ALP + B-SR & 207m\\[0.3em]
        \multirow{9}{*}{\snips} & BERT IC-SL & 23m\\
        & + B-SR & 29m \\
        & + Uni A-DA & 37m \\
        & + Uni A-DA + B-SR & 42m \\
        & + Uni A-DA + ALP & 59m \\
        & + Uni A-DA + ALP + B-SR & 70m \\
        & + BP A-DA & 102m \\
        & + BP A-DA + B-SR & 110m \\
        & + BP A-DA + ALP & 186m \\
        & + BP A-DA + ALP + B-SR & 227m \\
        \bottomrule
    \end{tabular}
    \caption{Table showing the time taken to run 30 epochs of a model training + inference on test data. As the test data size varies based on the noise type (since different treatment sets have different number of utterances), all numbers are shown for the abbreviation control set.}
    \label{tab:times}
\end{center}
\end{table}

\subsection{Evaluation Metrics}

We compute slot labeling F1 score using the seqeval library (\url{ https://github.com/chakki-works/seqeval}). We utilize the evaluation function provided in the gluonnlp intent classification and slot labeling tutorial to compute intent classification accuracy (\url{https://nlp.gluon.ai/model_zoo/intent_cls_slot_labeling/index.html}).

\begin{table*}[t]
\begin{center}
\begin{tabular}{ l c l l}
Noise Type & Probability & Example & Char. Edit\\
\hline
Original      & 1.0 - $p$    & list \textit{flights} from las vegas to phoenix & N/A \\
Insertion     & $p * 0.33$   & list \textit{fljights} from las vegas to phoenix & $+$j\\
Deletion      & $p * 0.18$   & list \textit{flghts} from las vegas to phoenix & $-$i\\
Substitution  & $p * 0.43$   & list \textit{flithts} from las vegas to phoenix & g $\rightarrow$ t \\
Transposition & $p * 0.06$   & list \textit{filghts} from las vegas to phoenix & l $\xleftrightarrow{}$ i \\
\end{tabular}
\caption{Instances of insertion, deletion, substitution, and transposition noise types for the example utterance (Original), ``list flights from las vegas to phoenix". We sample each noise type with the given probability to construct noisy pre-training and IC/SL training datasets, where $p$ is the noise rate. We inject noise into the token \textit{``flights"} in each example and provide the character level edit  (Char. Edit) that transforms the original token to the noised token. }
\label{tab:typos}
\end{center}
\end{table*}

\subsection{Model Training and Inference Times}
Training for $40$ epochs plus inference on the test set takes approximately 13 minutes on the \atis data-set and 23 minutes for the \snips data-set for both the baseline and training augmentation approaches. The time taken significantly increased for the other model types and are listed in \autoref{tab:times}.

\subsection{Synthetic Misspelling Generation}

We build on prior work by \cite{hasan-etal-2015-spelling} to generate synthetic misspellings that are representative of natural misspellings. Hasan et al. present a taxonomy of misspelling types and induction probabilities mined from natural noise. This taxonomy consists of four noise types, substitution, insertion, deletion, and transposition. 

\newpage

Character choice in substitution and insertion operations is based on the QWERTY keyboard layout. Given a character $c$ (e.g., "d"), we substitute or insert a character the appears next to $c$ on the QWERTY keyboard (e.g. "f", "s", "e", "c"). We provide examples of each noise type and list the probability with which we introduce these types of noise in \autoref{tab:typos}.

\end{document}